\RequirePackage{rotating}

\documentclass[sigconf,10pt]{acmart}
\makeatletter
\def\@ACM@checkaffil{
    \if@ACM@instpresent\else
    \ClassWarningNoLine{\@classname}{No institution present for an affiliation}%
    \fi
    \if@ACM@citypresent\else
    \ClassWarningNoLine{\@classname}{No city present for an affiliation}%
    \fi
    \if@ACM@countrypresent\else
        \ClassWarningNoLine{\@classname}{No country present for an affiliation}%
    \fi
}
\makeatother

\usepackage{multirow}
\usepackage{float}
\usepackage{makecell}
\usepackage[graphicx]{realboxes}
\usepackage{makecell}
\usepackage{booktabs}

\usepackage{amssymb}
\usepackage{mathtools}
\usepackage{lipsum}
\usepackage{pifont}
\usepackage{graphicx}
\usepackage{gensymb}
\usepackage{amssymb}
\usepackage{mhchem}
\usepackage{lipsum}
\usepackage{caption}
\usepackage{pdflscape} 
\usepackage{rotating}
\usepackage{enumitem}
\usepackage{titlesec}
\usepackage{graphicx}
\usepackage[bookmarksnumbered,unicode]{hyperref}
\usepackage{hyperxmp}
\usepackage{textpos} 

\usepackage{listings}
\titlespacing*{\section}{0pt}{1.3ex}{1.3ex}
\titlespacing*{\subsection}{0pt}{0.5ex}{0.5ex}
\titlespacing*{\subsubsection}{0pt}{0.5ex}{0.5ex}
\DeclareUnicodeCharacter{2212}{-}

\AtBeginDocument{%
  }

\copyrightyear{2025}
\acmYear{2025}
\setcopyright{cc}
\setcctype{by}
\acmConference[ACM MOBICOM '25]{The 31st Annual International Conference on Mobile Computing and Networking}{November 4--8, 2025}{Hong Kong, China}
\acmBooktitle{The 31st Annual International Conference on Mobile Computing and Networking (ACM MOBICOM '25), November 4--8, 2025, Hong Kong, China}\acmDOI{10.1145/3680207.3765264}
\acmISBN{979-8-4007-1129-9/2025/11}

\begin{document}

\title{Benchmarking Ultra-Low-Power $\mu$NPUs}


\author{Josh Millar}
\orcid{0009-0002-2247-1594}
\affiliation{
 \institution{Imperial College London}
 \city{}
 \country{}
}

\author{Yushan Huang}
\orcid{009-0009-4570-9455}
\affiliation{
 \institution{Imperial College London}
 \city{}
 \country{}
}

\author{Sarab Sethi}
\orcid{0000-0002-5939-0432}
\affiliation{
 \institution{Imperial College London}
 \city{}
 \country{}
}

\author{Hamed Haddadi}
\orcid{0000-0002-5895-8903}
\affiliation{
 \institution{Imperial College London}
 \city{}
 \country{}
}

\author{Anil Madhavapeddy}
\orcid{0000-0001-8954-2428}
\affiliation{
 \institution{University of Cambridge}
 \city{}
 \country{}
}

\renewcommand{\shortauthors}{Millar et al.}

\begin{abstract}

Efficient on-device neural network (NN) inference offers predictable latency, improved privacy and reliability, and lower operating costs for vendors than cloud-based inference. This has sparked recent development of microcontroller-scale NN accelerators, also known as neural processing units ($\mu$NPUs), designed specifically for ultra-low-power applications.

We present the first comparative evaluation of a number of commercially-available $\mu$NPUs, including the first independent benchmarks for multiple platforms. To ensure fairness, we develop and open-source a model compilation pipeline supporting consistent benchmarking of quantized models across diverse microcontroller hardware. 
Our resulting analysis uncovers both expected performance trends as well as surprising disparities between hardware specifications and actual performance, including certain $\mu$NPUs exhibiting unexpected scaling behaviors with model complexity. This work provides a foundation for ongoing evaluation of $\mu$NPU platforms, alongside offering practical insights for both hardware and software developers in this rapidly evolving space.
\end{abstract}



\maketitle


\vspace{-3mm}
\section{INTRODUCTION}

Performing neural network (NN) inference on constrained devices has applications across numerous domains, including wearable health monitoring \cite{wearablehealth}, smart agriculture \cite{agriculture}, real-time \emph{earable} audio processing \cite{audio}, and predictive maintenance \cite{predmaintenance}. Embedding inference also offers improved privacy over cloud-based alternatives, by eliminating the need to transmit sensitive data, alongside improved latency for time-critical applications, reduced operating costs for vendors, and improved reliability by removing dependence on network connectivity. Given their unique form factor and low power consumption, microcontrollers (MCUs) are widely used in resource-constrained environments. However, their performance is also often constrained by limitations in memory capacity, throughput, and compute.
The computational demands of modern neural networks (NNs) have catalyzed the development of specialized hardware accelerators across the computing spectrum, from high-\\performance data centers to ultra-low-power and embedded devices. At the resource-constrained end of the spectrum, microcontroller-scale neural processing units ($\mu$NPUs) have recently emerged, designed to operate within extremely tight power envelopes -- in the milliwatt or sub-milliwatt range -- while still providing latency sufficient to support real-time inference. These platforms represent a new class of accelerator, combining the power efficiency of MCUs with the cognitive capabilities previously exclusive to more powerful computing platforms. The core advantage of $\mu$NPUs stems from their ability to exploit the inherent parallelism of neural networks with dedicated multiply-accumulate (MAC) arrays alongside specialized memory structures for weight storage. This architectural specialization enables $\mu$NPUs to achieve orders of magnitude improvement in latency compared to general-purpose MCUs executing equivalent workloads.

Despite the growing availability of $\mu$NPU platforms, the field lacks a standardized and comprehensive evaluation or benchmark suite. Existing benchmarks focus solely on Analog Devices' MAX78000 \cite{max78000, moss2022ultra, clay2022benchmarking}, lacking side-by-side comparison with other platforms. Naturally, hardware vendors themselves provide performance metrics, but these are usually based on proprietary evaluation frameworks, using disparate NN models, quantization strategies, and other varying optimizations. This heterogeneity across evaluation methods, and absence of independent verification of vendor-provided performance claims, creates uncertainty for hardware designers and embedded software developers in selecting the most suitable platform for their application’s constraints. 
The lack of standardized benchmarks also hampers research by obscuring the relationship between architectural design and inference performance. Given the rapid pace of development and increasing diversity of available $\mu$NPU platforms, establishing reliable comparative benchmarks has become an urgent need. 

\noindent To this end, we make the following contributions:
\begin{itemize}[topsep=1mm, itemsep=0mm, parsep=0mm, left=0mm] 
\item \textbf{Side-by-Side Benchmark of $\mu$NPU Platforms:} We conduct the first comparative evaluation of commercially-available $\mu$NPU platforms, enabling direct performance comparisons across diverse hardware architectures under consistent workloads and measurement conditions.
\item \textbf{Independent Platform Benchmarks:} We also provide the first fine-grained and independent performance benchmarks for several $\mu$NPU platforms that have not previously been subject to third-party evaluation, offering unbiased verification of vendor performance claims.
\item \textbf{Open-Source Model Compilation Toolchain:} We develop and release\footnote{\url{ https://github.com/j0shmillar/uNPU-Bench}} an open-source toolchain to support consistent and simplified transplanting of NN models across our $\mu$NPU platforms, reducing the engineering overhead associated with cross-platform evaluation.
\item \textbf{Recommendations for Developers:} Informed by our benchmark results, we provide actionable recommendations to developers regarding platform selection, key focus areas for model optimization, and trade-offs for various applications and constraints.
\end{itemize}

In developing a unified compilation and benchmarking pipeline, we standardize model representations across the various $\mu$NPU platforms, enabling direct comparison of latency, memory, and energy performance. The evaluation also includes fine-grained analysis of the various stages of model execution, from NPU initialization and memory input/output overheads to CPU pre/post-processing -- aspects that can significantly impact end-to-end performance but are often not addressed in technical evaluations. The resulting analysis uncovers both expected performance trends as well as surprising disparities between hardware specifications and actual performance, including certain $\mu$NPUs exhibiting unexpected scaling behaviors with increasing model complexity. We hope our findings provide valuable insights to both developers and hardware architects.

 
\section{BACKGROUND \& MOTIVATION}
\subsection{Constrained Neural Computing}
The shift from cloud-based to on-device neural computing has numerous advantages for real-time data processing, especially with increasing concerns regarding data privacy and security \cite{kong2022edge}. Unlike cloud-based solutions, local inference mitigates security risks by processing sensitive data locally, which is particularly advantageous in domains such as medical diagnostics and surveillance \cite{wang2022privacy, wang2023security}. Additionally, local processing reduces end-to-end latency alongside operating costs for model vendors. However, traditional NN accelerators, such as GPUs and TPUs, are ill-suited to resource-constrained environments given their power consumption and large form factors \cite{kim2023hardware, lin2020mcunet}.

MCUs are compact, low-power computing platforms, often reliant on a single CPU and shared memory bus \cite{saha2022machine}. While MCUs are commonly adopted for resource-constrained IoT applications \cite{lin2022device, kwon2023tinytrain, huang2024low}, they generally lack the computational resources for efficient NN inference. Specifically, the computational capability of typical MCUs is often limited to a few million MAC operations per second, far below the tens of billions MACs/s required for real-time NN inference. Their absence of dedicated hardware acceleration results in large latency overheads and elevated power consumption during NN processing. Limited SRAM and flash memory also often poses challenges for efficiently managing the large weight matrices required for NN inference.

Given the various shortcomings of traditional MCUs, \\ microcontroller-scale \textit{$\mu$NPUs} are emerging as a response. These specialized NN accelerators offer dedicated neural processing hardware, providing higher throughput for NN workloads, meeting the stringent requirements of real-time NN inference \cite{manor2022custom, wang2022bed, song2024tada} while maintaining low-power operation. Collectively, $\mu$NPUs position themselves as a key solution for real-time NN processing in low-power environments.

\subsection{$\mu$NPU Hardware Design}

\begin{figure}[t!]
\begin{center}
\includegraphics[width=5cm]{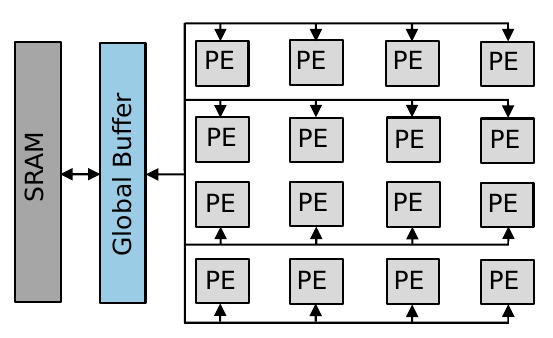}
\vspace{-0.3cm}
\caption{typical $\mu$NPU hardware architecture}\label{fig:arch}
\end{center}
\vspace{-0.6cm}
\end{figure}

$\mu$NPU hardware design is optimized for efficient tensor operations via specialized MAC units and parallelizable memory hierarchies \cite{caronti2023fine, gong2025dex}. Fig. \ref{fig:arch} illustrates the architecture of a typical $\mu$NPU, composed of a systolic array of processing elements (PEs). Notably, each PE contains its own MAC units and, importantly, its own weight memory space to avoid memory contention and maximize parallelization. The array of PEs is linked by an inter-PE communication grid, which connects to a large global buffer and SRAM/DRAM via an on-chip network \cite{sze}. Efficient memory hierarchy optimization is achieved by partitioning available RAM, along with implementing high-bandwidth memory interfaces and data prefetching mechanisms, addressing the memory bottlenecks faced by traditional MCUs when handling large NN weights. $\mu$NPUs mainly vary by their number of PEs, PE layout and clustering, memory hierarchy layout, and the availability/amount of storage/MAC units in each PE. 

These architectural advantages, coupled with low-power optimization techniques such as 
power gating, enable $\mu$NPU platforms to deliver low-power, high-throughput performance for real-time NN inference.


\subsection{Benchmarking $\mu$NPU Platforms}


\noindent \textbf{Need for Comprehensive Benchmarking:} Existing work on $\mu$NPU platforms mainly focuses on practical applications and/or model optimizations 
\cite{giordano2021battery, bakar2022protean, yolo_squeeze}, lacking fine-grained performance analysis from a systems perspective. In evaluating memory usage, latency, power, and throughput, across $\mu$NPU platforms, we aim to uncover critical performance bottlenecks, guiding researchers towards more efficient software and NN model design. 

\noindent \textbf{Limitations of Existing Benchmarks:} Existing benchmarks of $\mu$NPUs focus on a single platform, lacking horizontal comparisons across the now wide variety of available platforms \cite{moss2022ultra, clay2022benchmarking, huang2024energy}. This narrow perspective limits understanding of the variations in performance and applicability across different $\mu$NPUs. Existing standalone benchmarks also have significant shortcomings. Chiefly, most focus solely on the model's inference forward pass, overlooking other adjacent operations within the end-to-end model inference or application pipeline(s), such as NPU initialization, memory input/output (I/O), and CPU pre/post-processing. While often neglected, these factors can significantly impact overall 
performance and efficiency.
 
\section{INFRASTRUCTURE \& METHODOLOGY}

\vspace{-0.2cm}
\subsection{Hardware}
\begin{figure}[t!]
\begin{center}
\includegraphics[width=6cm]{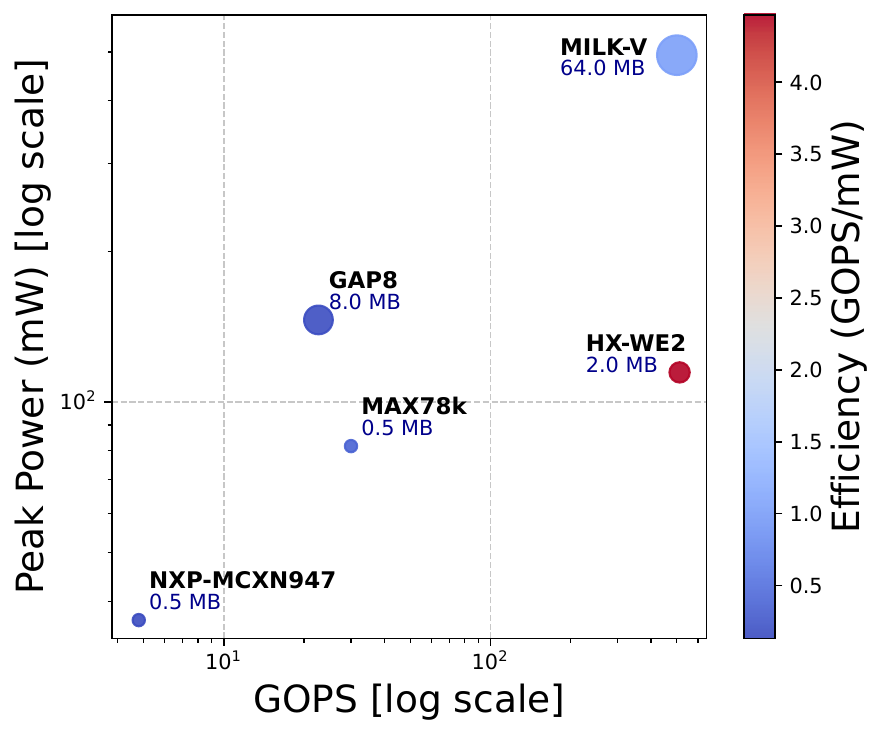}
\vspace{-0.3cm}
\caption{the various $\mu$NPUs used in our benchmark, and how they compare in terms of max GOPS, peak power draw, and \textit{theoretical} efficiency (GOPS/mW).}\label{fig:gops_vs_pp_log}
\end{center}
\vspace{-0.5cm}
\end{figure}

To provide a comprehensive benchmark, we evaluate a diverse range of widely-used, commercially-available $\mu$NPU platforms, from ultra-low-power $\mu$NPUs to NPU-equipped system-on-chip (SoC) architectures. These are evaluated alongside MCUs without dedicated neural hardware for comparison. Our selection covers a wide range of computational capabilities (<5 to >500 GOPs), memory configurations (128 KB to 2 MB RAM), and bit-width support (1-bit quantized to 32-bit floating-point operations). Fig. \ref{fig:gops_vs_pp_log} provides a visualization of peak GOPs (Giga Operations Per Second) vs. peak power for the various $\mu$NPU platforms included in our benchmark (on a log scale).
Table \ref{tab:boards} details our set of benchmark $\mu$NPUs, and we provide more detail on each platform below. 

The \textbf{MAX78000} (or \textbf{MAX78K}) \cite{max78000} from US-based Analog Devices features a Cortex-M4F with a RISC-V coprocessor, each capable of acting as the primary processor, along with a proprietary 30-GOPS CNN accelerator. The latter has a dedicated 512 KB SRAM for input data, 442 KB for weights, and 2 KB for biases, and supports quantized operations at 1, 2, 4, and 8-bit precision. The same fine-grained bit-width quantization is not yet widely supported on other $\mu$NPU platforms, or indeed in common software libraries designed for ML on resource-constrained devices; TFLite/LiteRT \cite{tflm}, for example, only supports 8-bit integer and 16-bit float weight quantization. The MAX78000 also has 512 KB of flash and 128 KB of CPU-only SRAM. This platform is among the best-documented commercially-available $\mu$NPUs; previous work has benchmarked its CNN accelerator under various configurations \cite{moss2022ultra, clay2022benchmarking, huang2024energy}, alongside exploring optimal model and data loading strategies for its 2D memory layout \cite{tinymem}. 

The \textbf{GAP8} \cite{gap8}, part of GreenWaves Technologies’ GreenWaves Application Processor series, features an 8-core RISC-V cluster and  22.65-GOPS hardware convolution engine for neural network acceleration at 8 or 16-bit precision.  The platform has 512 KB of L2 RAM, up to 8 MB of L3 SRAM, and 20MB flash storage, enabling it to store and run larger, more complex models or mixture-of-experts (MoE) architectures. The GAP series of $\mu$NPUs have also been the subject of several recent works, again mainly centered on model optimization \cite{yolo_squeeze, gap81, gap91}; no platform benchmark exists yet.

The \textbf{Himax HX6538 WE2} (or HX-WE2) \cite{himax} is a more powerful $\mu$NPU platform from Taiwan-based semiconductor manufacturer, Himax Technologies. This platform features a Corstone-300 set up, with Cortex M55 CPU and Ethos U55 NPU, delivering up to 512 GOPS.  The platform also features 512KB TCM, 2MB SRAM, and 16MB flash, suitable for large or more complex models, but at an increased power draw. 

NXP's \textbf{MCXN947} \cite{mcxn947} is part of the MCX N94x line of MCUs, featuring dual Cortex-M33 CPUs and NXP's eiQ Neutron NPU. The MCXN947 is designed for lower-power applications, with 8-bit neural acceleration of only 4.8 GOPS. The platform features 512 KB RAM and 2 MB flash storage.

Our benchmark also includes MCUs without neural hardware for comparison, to quantify any efficiencies gained from specialized NPU architectures. 

The \textbf{STM32H7A3ZI} \cite{stm32} is a high-performance MCU based on the Cortex M7, with 2 MB of flash and 1.4 MB of SRAM. Manufactured by Swiss-based ST Microelectronics, it is frequently used with on-board NNs \cite{addabbo2022smart,huang2024low}. 

The \textbf{ESP32s3} MCU \cite{esp32} features dual-core Tensilica LX6 processors, 512 KB of SRAM, 2MB PSRAM, and 8MB flash. Notably, whilst primarily a low-power MCU, it advertises NN acceleration capabilities with “support for vector instructions ... providing acceleration for neural network computing". This is achieved via an extended instruction set, which includes 128-bit vector operations, $e.g.$, complex multiplication, addition, subtraction, shifting, and comparison.

We also include the \textbf{MILK-V Duo} \cite{milkv}, a RISC-V SoC built around the CVITEK CV1800B processor. Unlike the previous MCUs/$\mu$NPUs, it can run Linux variants or an RTOS, such as FreeRTOS, supporting more flexible NN workloads at a much-increased power budget. This platform represents the upper bound of our evaluation in terms of computational power and software flexibility.

\begin{table}[htbp!]
    \vspace{-0.2cm}
    \centering
    \caption{the $\mu$NPU platforms used in our benchmark.}
    \vspace{-0.1cm}
    \small
    \scriptsize 
    \setlength{\tabcolsep}{2pt} 
    \renewcommand{\arraystretch}{1.05} 
    \resizebox{\columnwidth}{!}{
    \begin{tabular}{lccccccc}
        \toprule
        \textbf{MCU} & \textbf{CPU(s)} & \textbf{NPU}  & \textbf{Flash} & \textbf{RAM} & \textbf{GOPs (max)} & \textbf{Bit Cap.}\\ [-1mm]
        \midrule
         \makecell[l]{MAX78000} &  \makecell[c]{Cortex-M4 \\ RISC-V} & MAXIM-own & 512KB & \makecell[c]{512 KB NPU \\ 128 KB CPU} & 30 & 1, 2, 4, 8  \\ [-3mm]
         
         \makecell[l]{HX-WE2 \\ (Corstone-300)} & Cortex-M55 & Ethos-U55 & 16 MB & \makecell[c]{2 MB SRAM \\ 512 KB TCM} & 512 & 8,16,32\\ [-3mm]
         
         \makecell[l]{NXP-MCXN947} & Cortex-M33 (2) & eIQ Neutron & 2 MB & 512 KB & 4.8 & 8 \\ [-3mm]
         
         \makecell[l]{GAP8} & RISC-V &  GAP-own & 20 MB L3 & \makecell[c]{512 KB L2 \\ 8MB L3} &  22.65 & 8,16 \\ [-2mm]

         \makecell[l]{STM32H74A3ZI} & Cortex-M7 & - & 2 MB & 1.4 MB & - & 8, 16, 32 \\ [-3mm]

         \makecell[l]{ESP32} & Tensilica LX6 & - & 4 MB & 520 KB & - & 8, 16, 32 \\ [-2mm]

         \makecell[l]{MILK-V} & \makecell[c]{RISC-V \\ XuanTie C906 (2)} & CV1800B & - & 64 MB & 500 & 8, 16, 32 \\ [-2mm]
         
        \bottomrule
    \end{tabular}
    }
    \vspace{-0.2cm}
\end{table}
\label{tab:boards}

\subsubsection{Note on CPU Frequency} 
We configure the various $\mu$NPU platforms to operate at a uniform CPU frequency. While this permits direct comparison of architectural efficiency, it should be noted that many of the platforms are capable of operating at higher frequencies than evaluated -- approaching the GHz range in some cases. Our method intentionally isolates architectural efficiency, but further experimentation could explore the impact of varying CPU frequencies on end-to-end latency and power consumption.
All other hardware parameters are fixed to platform-default settings to ensure comparability. These include the number of active processing elements (PEs) or compute cores, on-chip SRAM banking configuration, and any vendor-specific accelerator tuning parameters ($e.g.$, tiling sizes, DMA burst length, cache enablement). Memory layout settings -- such as channel ordering -- are likewise determined by the compiler defaults for each platform. While these defaults yield a ``fair" cross-platform baseline, they may not represent peak achievable performance.

\subsection{Models}
Table \ref{tab:model_intro} details the various CNN-based models used in our benchmark, covering image classification, object recognition, and signal reconstruction applications. We provide more detail on each model below.

\noindent\textbf{CIFAR10-NAS:} the optimal memory-constrained CNN model for the CIFAR-10 dataset, generated using the Once-for-All (OFA) NAS framework, a weight-sharing-based framework that decouples search and training by constructing a \textit{supernet} model from which various hardware-specific \textit{subnet} models can be derived \cite{onceforall}. . It combines diverse convolutional units, frequent 1×1 convolutions for channel mixing, alternating pooling layers, and irregular channel scaling -- patterns that are characteristic of hardware-aware NAS designs. This is our largest model, with 74.3 Million MACs (MMACs) and 36.4 Million FLOPs (MFLOPs). Trained on the CIFAR-10 dataset, with 3x32x32 input size and 10-class output. 

\noindent\textbf{SimpleNet:} a simpler CNN framework composed of a basic stack of convolutional and pooling layers \cite{simplenet}.
SimpleNet has 38.0 MMACs and 18.5 MFLOPs. Trained on the CIFAR-100 dataset, with 3x32x32 input size and 100-class output.

\noindent\textbf{ResidualNet:} a SimpleNet variant built around residual functions, helping to mitigate gradient vanishing and introducing a non-trivial scheduling problem for inference compilers. ResidualNet and SimpleNet are structurally matched in size and compute, but differ in connectivity, enabling direct measurement of skip-connection overheads. ResidualNet has 37.7 MMACs and 18.5 MFLOPs. Trained on the CIFAR-100 dataset, with 3x32x32 input size and 100-class output.

\noindent\textbf{AI8XAutoEnc:} encoder–decoder architecture adapted from the AI8X framework (\textit{i.e.}, for MAX78000/2). It combines two initial 1D convolutional layers with a series of fully connected layers, including a strong bottleneck compression down to 4 latent features before reconstruction. The decoder uses small FC expansions rather than mirroring the encoder’s convolutional structure. This model is our simplest, with just 0.5 MMACs and 0.2 MFLOPs. Trained on a machine fault detection dataset, generated using the SpectraQuest Machinery Fault Simulator \cite{MotorDatasets}. The input/output size is 3x256.

\noindent\textbf{YoloV1:} a compact, single-stage object detection CNN, with its final layers pruned for uniformity across platforms. This network is deep (20 convolutional layers) but narrow (maximum 32 channels), with heavy use of 1×1 reductions and rapid spatial downsampling via max pooling. YoloV1 has 43.83 MMACs and 21.2 MFLOPs. Trained for person-only detection on the COCO dataset \cite{coco}, with input size 3x96x96. The network produces multi-scale output feature maps for bounding boxes and class probabilities, requiring CPU-side non-max suppression (NMS).


\begin{table*}[t!]
    \setlength{\tabcolsep}{3pt}
    \centering
    \caption{The various models used in our benchmark. Note: MACs/FLOPs are forward-only. Peak activation RAM and Lifetime Pressure are measured with batch=1 and INT8 activations.}
    \vspace{-0.1cm}
    \resizebox{\textwidth}{!}{
    \begin{tabular}{lccccccccccc}
        \toprule
        \textbf{Model} &
        \textbf{Input} &
        \textbf{Output} &
        \textbf{\#Params} &
        \textbf{\#Layers} &
        \textbf{Conv Depth} &
        \textbf{Max Conv Ch.} &
        \textbf{FC Depth} &
        \textbf{Peak Act. RAM (MB, B=1)$^{\dagger}$} &
        \textbf{Lifetime Pressure (MB$\cdot$steps)$^{\ddagger}$} &
        \textbf{MMACs} &
        \textbf{MFLOPs} \\
        \midrule
        \textbf{CIFAR10\text{-}NAS} & 3$\times$32$\times$32 & 1$\times$10  & 298{,}762 & 11 & 10 & 128 & 1 & 0.07 & 18.39 & 74.25 & 36.38 \\
        \textbf{ResidualNet}        & 3$\times$32$\times$32 & 1$\times$100 & 383{,}012 & 14 & 14 & 512 & 0 & 0.14 & 2.98  & 37.78 & 18.46 \\
        \textbf{SimpleNet}          & 3$\times$32$\times$32 & 1$\times$100 & 383{,}012 & 14 & 14 & 512 & 0 & 0.14 & 2.45  & 38.00 & 18.46 \\
        \textbf{AI8XAutoEnc}        & 3$\times$256          & 3$\times$256 & 136{,}800 & 7  & 2  & 128 & 5 & 0.07 & 2.89  & 0.55  & 0.20  \\
        \textbf{YoloV1}             & 3$\times$96$\times$96 & \makecell[c]{1$\times$12$\times$12$\times$12\\1$\times$12$\times$12$\times$2\\1$\times$12$\times$12$\times$10} & 40{,}700 & 20 & 20 & 32  & 0 & 0.02 & 0.80  & 43.83 & 21.22 \\
        \bottomrule
    \end{tabular}}
    \footnotesize
    \begin{flushleft}
    $^{\dagger}$ Peak activation RAM: maximum live activation footprint during the forward pass (batch=1, INT8), excluding parameter storage.\\
    $^{\ddagger}$ Lifetime Pressure: $\sum_{t} \big(\text{size}(t)\_\mathrm{MB} \times \text{lifetime\_steps}(t)\big)$ across all tracked activations, where $\text{lifetime\_steps}(t)$ counts the number of subsequent leaf-module steps over which activation $t$ must remain live until its last use. Higher values indicate reduced opportunities for buffer reuse.
    \end{flushleft}
    \label{tab:model_intro}
    \vspace{-0.4cm}
\end{table*}

\vspace{-0.1cm}


\subsubsection{Ensuring Model Uniformity} 


We encountered substantial variability in operator support across the benchmark platforms. The NXP-MCXN947's eIQ Neutron NPU lacks native support for softmax operations, for example, necessitating its implementation as a CPU post-processing step for relevant models. Similarly, operations associated with non-maximum supression (NMS) in the YoloV1 model were inconsistently supported across platforms, requiring us to also move the entire NMS operation to CPU post-processing. This explains the unusual multi-component output shape of our YoloV1 model (see Table \ref{tab:model_intro}). The benchmark platforms also outline varying levels of support for operator compatibility. The MAX78000, for example, only supports 1D convolution with kernel sizes 1 to 9 and 2D convolution with kernel sizes of 1 by 1 or 3 by 3. Unsupported operations will fall back to CPU execution and incur latency penalties.

By identifying and constructing models using a core subset of operators that are universally supported across all $\mu$NPUs, we aim to ensure that any measured performance differences stem from fundamental architectural discrepancies rather than variations in model compilation and optimization.

\subsubsection{Quantization}

We quantize all benchmark models to \texttt{INT8} precision, as it is supported by all evaluated NPUs. However, it's important to note that while this enables a more direct architectural comparison, it may not reflect the optimal accuracy-performance tradeoff on each platform; some NPUs, such as the MAX78000, support lower bit-widths ($e.g.$, 1, 2, 4-bit), and others, like the HX-WE2 support floating-point acceleration ($e.g.$, FLOAT16 and 32-bit).

We perform post-training quantization (PTQ) on all models/platforms. While platforms like the MAX78000 support quantization-aware training (QAT) and fused operators, such optimizations produce platform-specific models incompatible with other NPUs. PTQ enables us to maintain structural consistency across all platforms. Moreover, since our primary metrics of interest are latency and power consumption, rather than inference accuracy, PTQ provides a sufficiently representative model for performance evaluation. PTQ was performed using a representative calibration dataset appropriate to each model's domain. We did not apply per-channel quantization for weights, instead using per-tensor quantization to ensure compatibility across all platforms.

\subsubsection{Compilation} 
\label{section:compilation} 

The various $\mu$NPUs support a wide range of model formats, from platform-optimized versions of common model formats ($e.g.$, TFLite) to platform-specific formats ($e.g.$, CVITEK's CVIMODEL). To facilitate cross-platform deployment, we developed a custom model compilation workflow for converting our base models into optimized formats for each target NPU. Our workflow ingests Torch (or ONNX/TFLM) base models along with various compiler flags ($i.e.$, target NPU platform, model input dimensions, bit-precision requirements, representative PTQ calibration data, $etc$), producing platform-specific optimized models with accompanying inference code.

The compilation process varies significantly by platform. For example, models targeting the ARM Ethos-U55 (on the HX-WE2) are compiled using the ARM Vela compiler, which ingests TFLiteMicro (TFLM) models and produces binaries optimized for the Ethos-U architecture. Vela applies platform-specific optimizations, including memory reduction. We evaluate both the \textit{Size} optimization strategy, HX-WE2 (S), which minimizes SRAM usage, and the \textit{Performance} strategy, HX-WE2 (P), which prioritizes execution speed using available arena cache if specified.

For other platforms, we utilize their respective toolchains ($e.g.$, the MAX78k's SDK or the NXP eIQ portal tools). In each case, we configured such tools to maintain model structure equivalence while applying platform-appropriate optimizations. Note that in our model compilation workflow, template inference code is often generated along with a compiled model. However, this doesn’t include model-specific pre/post-processing steps, which should be implemented manually by the developer, who can update the template code as needed.

\begin{figure*}[t!]
\begin{center}
\includegraphics[width=16cm]{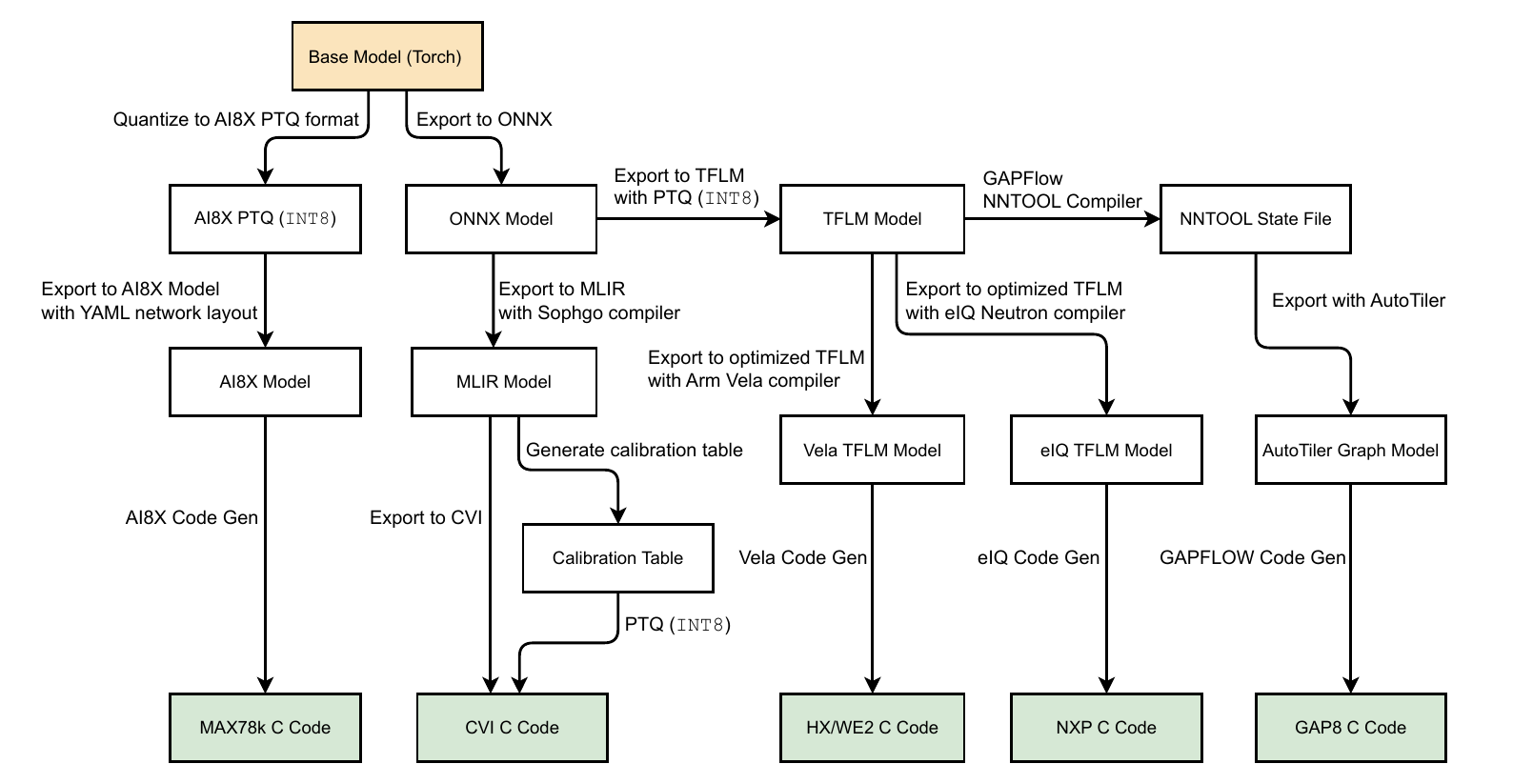}
\vspace{-0.3cm}
\caption{an overview of our model compilation workflow.}\label{fig:pipeline}
\end{center}
\vspace{-6mm}
\end{figure*}

Fig \ref{fig:pipeline} below details our model compilation toolchain for converting a base (Torch/ONNX/TFLM) model into various platform-specific formats. We open-source our toolchain\footnote{\url{https://github.com/j0shmillar/uNPU-Bench}} and hope its use can ease the process of cross-platform model compilation and benchmarking.

\subsection{Evaluation Metrics}
We measure latency, power, energy-efficiency -- in terms of number of inference operations per mJ -- and memory usage across each benchmark platform and model. The impacts of various platform-specific model optimizations or compilation workflows on model accuracy are out of scope for our study. Latency can be considered proportional to throughput, since batching and other amortization techniques are not practical on $\mu$NPU platforms due to memory constraints.

\noindent\textbf{Measurement Environment}: All platforms are measured without concurrent workloads, and all non-essential background processes are disabled where appropriate ($i.e.$, on MILK-V). This minimizes interference during repeated runs. Models are compiled and deployed using each platform’s vendor-provided SDK or toolchain in its default configuration, unless otherwise stated. Exact toolchain versions can be found in our accompanying repository. 

\noindent\textbf{Latency:} Latency is measured using each platform’s internal timer. Notably, all MCUs, bar the MILK-V, are configured to run at 100 MHz. The MILK-V lacks support for fixed frequency scaling, only DVFS. However, latency is inversely proportional to CPU frequency, as described by $T = N/f$ (where $T$ denotes latency, $N$ the number of cycles required for a task, and $f$ the operating frequency). Accordingly, we normalized MILK-V’s latency to be comparable to performance under uniform frequency conditions. 

Each model was run for 10 consecutive inferences. We report mean latency and standard deviation to account for run-to-run variability. We observe higher variance on the MILK-V SoC, primarily from CMA activity and residual background services. To reduce noise, we increased the number of runs for latency measurement from 10 to 100; while some variability remains, this averaging ensures stable estimates. Other platforms show limited variance under bare-metal execution, so 10 runs are sufficient. 

\begin{figure}[t!]
\begin{center}
\includegraphics[width=7.6cm]{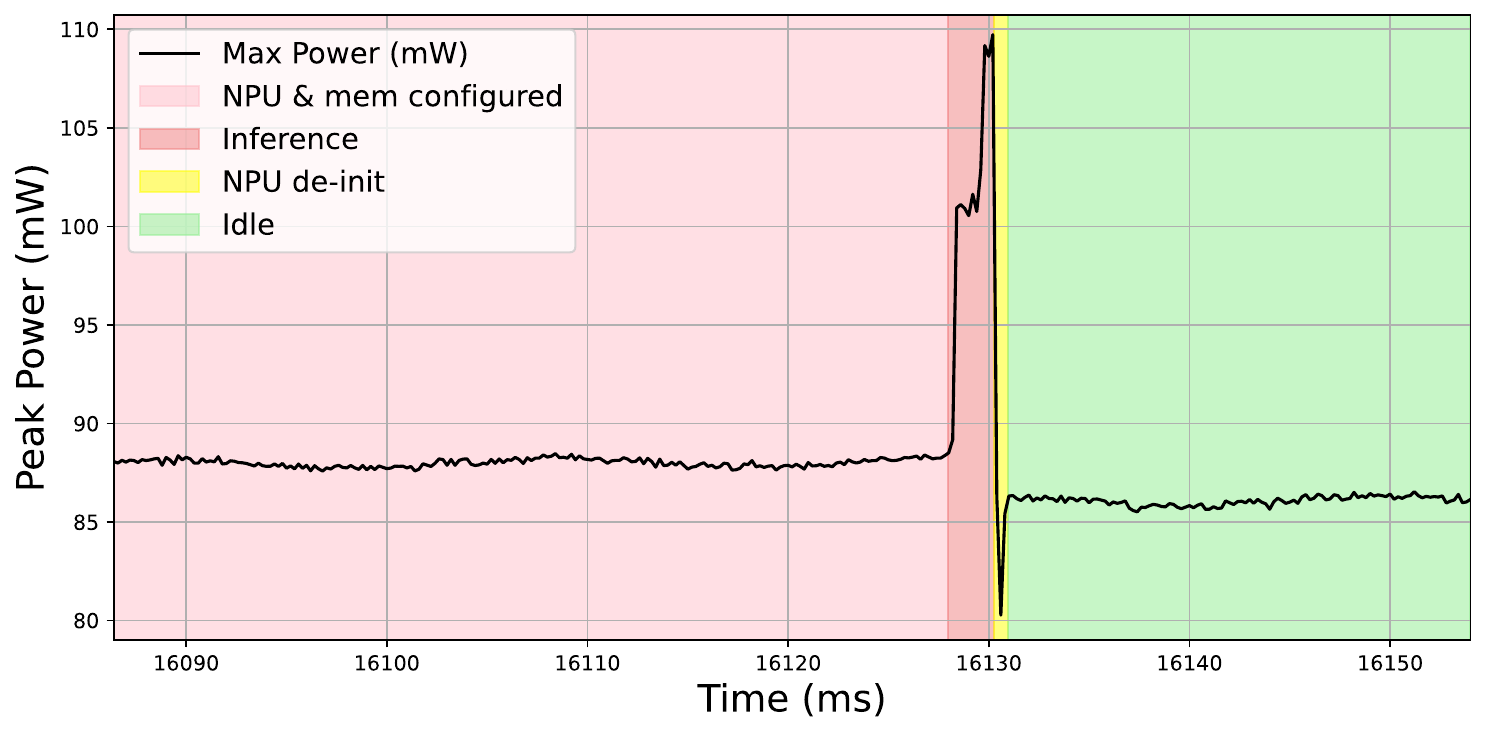}
\vspace{-0.3cm}
\caption{power trace of YoloV1 inference on HX-WE2.
}\label{fig:power_trace}
\vspace{-0.6cm}
\end{center}
\end{figure}

\noindent{\textbf{Power and Energy:}} We compute power and energy using a Monsoon High Voltage Power Monitor \cite{msoon}with PowerTool v5.0.0.10 software at a sampling rate of 50 kHz. The input voltage ($U$) is fixed at 3.3 V. We capture inference duration ($t$) and average power ($P$) to compute mean energy consumption ($E = P \cdot t$). To ensure steady-state measurements, readings are taken only after a 60 s warm-up period. Measurements are repeated for 10 inference runs, and mean ± standard deviation are reported. Fig. \ref{fig:power_trace} shows an example power profile for YOLOv1 inference on the HX-WE2’s Ethos-U55 $\mu$NPU.

\noindent{\textbf{Inferences per mJ:}} To quantify energy efficiency, we introduce `inferences per mJ', $I_{mJ}$, capturing the number of end-to-end inferences ($i.e.$, memory transfer, CPU pre/post-processing, and optionally NPU initialization) performed for each millijoule of energy consumed. 

\noindent{\textbf{Memory Usage:}} Memory usage is assessed by analyzing the linker ($.map$) file generated by the compilation toolchain. This file provides a detailed breakdown of memory allocation, including code ($.text$), initialized data ($.data$), and uninitialized data ($.bss$) segments. Flash memory usage is calculated as the sum of the code and initialized data segments ($.text + .data$), while RAM usage includes both the initialized and uninitialized data segments ($.data + .bss$). For the MAX78000, with its dedicated NPU-only memory, the RAM usage is computed separately for CPU and NPU.

\begin{figure}[htbp!]
\begin{center}
\vspace{-0.3cm}
\includegraphics[width=7cm]{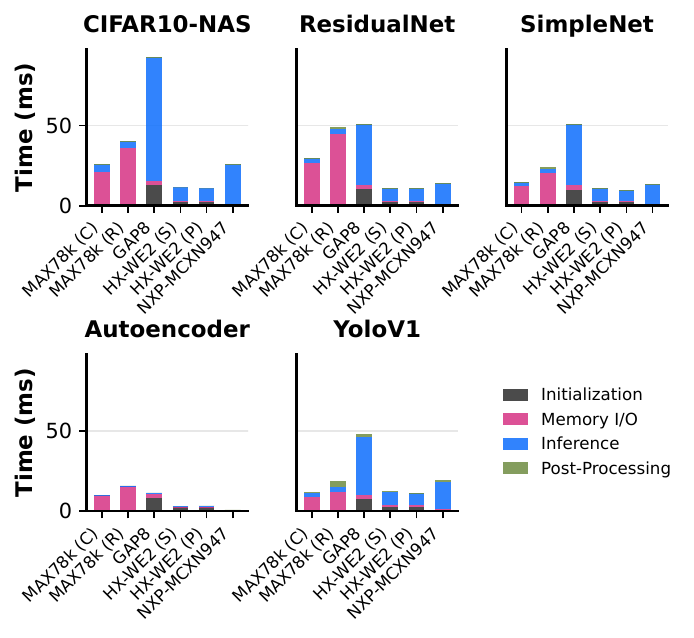}
\captionsetup{singlelinecheck=false}
\vspace{-0.4cm}
\caption{latency for each stage, model, and platform.\protect\footnotemark}
\label{fig:stacked_latency}
\end{center}
\vspace{-0.4cm}
\end{figure}

\footnotetext{MAX78k (C) denotes use of its Cortex-M CPU, and (R) its RISC-V CPU. HX-WE2 (S) denotes model compilation with the Vela \textit{Size} optimization flag, and (P) with the Vela \textit{Performance} flag.}
\subsection{Performance Breakdown}

We break down each stage of model execution and measure per-stage latency and power consumption. This granular analysis helps identify specific bottlenecks in the inference pipeline, alongside measuring overall end-to-end performance. We also measure idle power consumption ($i.e.$, each platform's base power draw in the absence of active computation). We provide more detail on each stage below:

\noindent\textbf{(NPU) Initialization:} This covers any NPU setup overhead, including memory buffer allocation and kernel configuration. 

\noindent\textbf{Memory I/O:} The cost/overhead of model and input data loading, including movement of input tensors and model weights from flash to NPU DRAM, and vice versa ($i.e.$, output tensors from NPU to CPU SRAM).

\noindent\textbf{Inference:} Executing the model's forward pass on the NPU. 

\noindent\textbf{Post-Processing:} Any operations required to be performed on the CPU. This includes computing softmax outputs for ResidualNet, SimpleNet, and CIFAR10-NAS models. YoloV1 post-processing includes NMS with output class softmax. 

\noindent\textbf{Idle:} The base power consumption of the various platforms, when not actively performing computation.

\noindent For MCUs without neural hardware ($i.e.$, the STM32H7A3ZI and ESP32), Initialization and Memory I/O are combined.

\begin{table*}[t!]
\centering
\caption{Inferences per mJ ($I_{mJ}$) for evaluated models and platforms, including NPU initialization. 
\\ The largest $I_{mJ}$ for each model is \underline{\textbf{underlined and bolded}}, while the second largest is \textbf{bold}.}
\label{tab:inf_per_j}
\vspace{-0.3cm}
\scriptsize
\setlength{\tabcolsep}{5pt}
\renewcommand{\arraystretch}{1.15}
\begin{tabular}{lccccccccc}
\toprule
 & \textbf{MAX78k (C)} & \textbf{MAX78k (R)} & \textbf{GAP8} & \textbf{NXP-MCXN947} & 
   \textbf{HX-WE2 (S)} & \textbf{HX-WE2 (P)} & \textbf{MILK-V} & \textbf{STM32H7A3ZI} & \textbf{ESP32s3} \\
\midrule
\textbf{CIFAR10-NAS} &
   \underline{\textbf{1.10}}{\scriptsize ±0.002} &
   0.85{\scriptsize ±0.001} &
   0.10{\scriptsize ±0.002} &
   \textbf{1.07}{\scriptsize ±0.002} &
   0.79{\scriptsize ±0.007} &
   0.83{\scriptsize ±0.006} &
   0.01{\scriptsize ±0.001} &
   0.03{\scriptsize ±0.001} &
   0.01{\scriptsize ±0.001} \\
\textbf{ResNet} &
   \textbf{1.24}{\scriptsize ±0.003} &
   0.85{\scriptsize ±0.002} & 
   0.17{\scriptsize ±0.002} &
   \underline{\textbf{1.97}}{\scriptsize ±0.003} &
   0.85{\scriptsize ±0.006} & 
   0.84{\scriptsize ±0.019} &
   0.01{\scriptsize ±0.001} & 
   0.06{\scriptsize ±0.001} & 
   0.02{\scriptsize ±0.001} \\
\textbf{SimpleNet} &
   \underline{\textbf{2.29}}{\scriptsize ±0.006} &
   1.65{\scriptsize ±0.003} &
   0.16{\scriptsize ±0.005} &
   \textbf{2.10}{\scriptsize ±0.004} &
   0.89{\scriptsize ±0.006} &
   0.99{\scriptsize ±0.006} &
   0.01{\scriptsize ±0.001} &
   0.07{\scriptsize ±0.001} &
   0.02{\scriptsize ±0.001} \\
\textbf{Autoenc} &
   \textbf{3.92}{\scriptsize ±0.014} &
   2.75{\scriptsize ±0.008} &
   1.12{\scriptsize ±0.028} &
   \underline{\textbf{36.95}}{\scriptsize ±0.002} &
   3.57{\scriptsize ±0.035} &
   3.06{\scriptsize ±0.038} &
   0.01{\scriptsize ±0.001} &
   3.48{\scriptsize ±0.082} &
   0.32{\scriptsize ±0.001} \\
\textbf{YoloV1} &
   \underline{\textbf{2.27}}{\scriptsize ±0.004} &
   1.76{\scriptsize ±0.003} &
   0.20{\scriptsize ±0.005} &
   \textbf{1.83}{\scriptsize ±0.006} &
   0.73{\scriptsize ±0.009} &
   0.81{\scriptsize ±0.008} &
   0.01{\scriptsize ±0.001} &
   0.05{\scriptsize ±0.001} &
   0.01{\scriptsize ±0.001} \\
\bottomrule
\end{tabular}
\end{table*}

\begin{table*}[htbp]
\centering
\caption{Inferences per mJ ($I_{mJ}$) for evaluated models and platforms, \emph{not} including NPU initialization. 
\\ The largest $I_{mJ}$ for each model is \underline{\textbf{underlined and bolded}}, while the second largest is \textbf{bold}.}
\label{tab:inf_per_j_no_init}
\vspace{-0.3cm}
\scriptsize
\setlength{\tabcolsep}{5pt}
\renewcommand{\arraystretch}{1.15}
\begin{tabular}{lccccccc}
\toprule
 & \textbf{MAX78k (C)} & \textbf{MAX78k (R)} & \textbf{GAP8} & \textbf{NXP-MCXN947} & 
   \textbf{HX-WE2 (S)} & \textbf{HX-WE2 (P)} & \textbf{MILK-V} \\
\midrule
\textbf{CIFAR10-NAS} &
   \textbf{1.11}{\scriptsize ±0.002} &
   0.85{\scriptsize ±0.001} &
   0.11{\scriptsize ±0.002} &
   1.09{\scriptsize ±0.002} &
   0.98{\scriptsize ±0.009} &
   1.04{\scriptsize ±0.008} &
   \underline{\textbf{2.80}}{\scriptsize ±0.077} \\
\textbf{ResNet} &
   1.24{\scriptsize ±0.003} &
   0.85{\scriptsize ±0.002} & 
   0.22{\scriptsize ±0.001} &
   \textbf{2.01}{\scriptsize ±0.002} &
   1.08{\scriptsize ±0.008} & 
   1.05{\scriptsize ±0.024} &
   \underline{\textbf{4.51}}{\scriptsize ±0.469} \\
\textbf{SimpleNet} &
   \textbf{2.30}{\scriptsize ±0.006} &
   1.66{\scriptsize ±0.003} &
   0.21{\scriptsize ±0.007} &
   2.13{\scriptsize ±0.004} &
   1.13{\scriptsize ±0.008} &
   1.29{\scriptsize ±0.010} &
   \underline{\textbf{4.17}}{\scriptsize ±0.195} \\
\textbf{Autoenc} &
   3.94{\scriptsize ±0.014} &
   2.78{\scriptsize ±0.008} &
   6.25{\scriptsize ±0.203} &
   \underline{\textbf{47.06}}{\scriptsize ±1.956} &
   \textbf{22.45}{\scriptsize ±0.392} &
   12.97{\scriptsize ±0.232} &
   13.29{\scriptsize ±0.883} \\
\textbf{YoloV1} &
   \textbf{2.27}{\scriptsize ±0.004} &
   1.76{\scriptsize ±0.003} &
   0.23{\scriptsize ±0.005} &
   1.86{\scriptsize ±0.007} &
   0.91{\scriptsize ±0.013} &
   1.03{\scriptsize ±0.011} &
   \underline{\textbf{5.75}}{\scriptsize ±0.274} \\
\bottomrule
\end{tabular}
\end{table*}

\section{RESULTS \& DISCUSSION}

Table \ref{tab:mega-comparison}, which can found in the supplementary material, details our full latency and power measurements across each stage, model, and platform.

\subsection{Power and Efficiency Breakdown}

Our results reveal significant variation in efficiency across the benchmark platforms, as shown in Tables ~\ref{tab:inf_per_j} and \ref{tab:inf_per_j_no_init}. 

The MAX78000 (C) with Cortex-M4 CPU active demonstrates the best \textit{overall} efficiency across evaluated models when NPU initialization overhead is considered, with consistent <30ms end-to-end latency. The MAX78000 (R) with RISC-V CPU lags slightly behind. This aligns with previous standalone benchmarks \cite{moss2022ultra}. 

The NXP-MCXN947 also achieves consistent sub-30ms latency, with its fast initialization and memory I/O offsetting the impact of (moderately) slower inference latency, delivering comparable (and in some cases, improved) efficiency despite its lower-throughput accelerator. 

Notably, the power-hungry but low-latency HX-WE2 platform, with Arm Corstone-300 (Cortex-M55 \& Ethos-U55 NPU), consistently beats the MAX78000 (C/R) in terms of end-to-end latency across the various models, due to the latter’s large memory I/O overhead. The HX-WE2 (S/P) demonstrates average $\sim$1.93x and $\sim$3.07x speedup in end-to-end latency over the MAX78000 (C) and (R) respectively, but $\sim$3.13x and $\sim$3.33x increase in average power consumption. We find the Vela \textit{Performance}-optimized models, for the HX-WE2, generally achieve slightly lower latency than the \textit{Size}-optimized models. However, their efficiency gain diminishes with model complexity -- efficiency on \textit{Performance}-optimized YoloV1 is lower than on its \textit{Size}-optimized variant.


The general-purpose MCUs without dedicated neural hardware -- the STM32H7A3ZI and ESP32s3 -- demonstrate significantly lower efficiency across all models. This result empirically validates the advantage neural hardware provides for performing on-device inference in constrained environments, with up to 2 orders of magnitude improvement in end-to-end latency in some cases. However, the STM32H7A3ZI's power consumption during inference (54.91 - 56.11 mW) is comparable to or lower than MAX78000 (C/R) for some models. This is particularly evident for the AI8XAutoEnc model, where the STM32H7A3ZI achieves a surprisingly competitive 3.483 $I_{mJ}$ -- comparable to the best-performing platforms in our suite. This is consistent with its architectural characteristics; it contains only two lightweight 1D convolutional layers followed by a predominantly fully-connected decoder, meaning its workload is compute-light and exhibits minimal parallelism for NPUs to exploit. As such, optimized scalar or SIMD execution on a high-performance MCU core ($e.g.$, STM32H7’s Cortex-M7) can achieve efficiency close to specialized hardware. In contrast, the ESP32 consistently exhibits high inference power consumption (129.74 - 157.17 mW) and latency (7.11 – 536.22 ms), despite its advertised support for CPU-accelerated tensor operations. Altogether, while general-purpose MCUs can achieve reasonable efficiency for simple models, they quickly become impractical for more complex NNs.

The MILK-V, our RISC-V SoC, also demonstrates low efficiency across all models, due to its NPU initialization overhead. We observe a different story, however, if initialization overhead is removed from consideration ($i.e.$, for continuous operation). Without initialization, the MILK-V ranks highest for efficiency across almost all benchmark models. Notably, despite a large idle power draw, it achieves blazingly fast inference times (0.17 - 0.61 ms). 

\begin{figure*}[t!]
\begin{center}
\includegraphics[width=\textwidth]{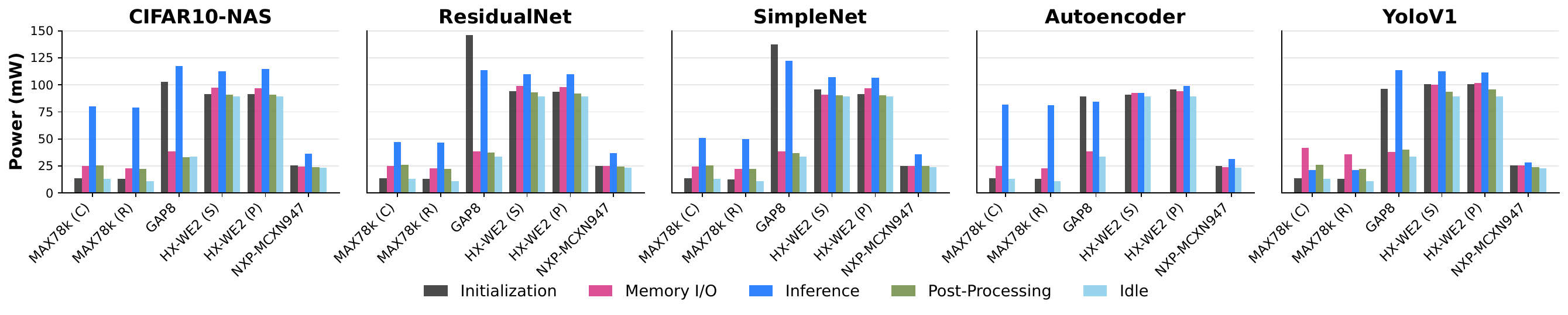}
\vspace{-7mm}
\caption{power consumption for each stage, model, and µNPU. }\label{fig:power}
\end{center}
\vspace{-3mm}
\end{figure*}

Fig. \ref{fig:power} details the power consumption breakdown across all evaluated platforms; among these, the MAX78000 (10.87–80.41 mW) and NXP-MCXN947 (22.91–36.69 mW) exhibit the lowest power draw across the benchmark models, with the NXP showing the lowest variance in peak power across the execution stages, enabling more reliable energy budgeting.

Beyond peak power, idle power consumption is another key consideration for low-power deployments, particularly if workloads run infrequently -- idle power also varies significantly across our benchmark platforms. The MAX78000 demonstrates the lowest idle power of the various $\mu$NPU platforms (10.87 mW with RISC-V and 13.21 mW under Cortex-M4). The HX-WE2 platform ranks highest (89.09 mW), raising concerns about its applicability in extremely power-constrained scenarios (such as ones in which long idle durations dominate overall energy usage).


\begin{figure}[t!]
\includegraphics[width=8.5cm]{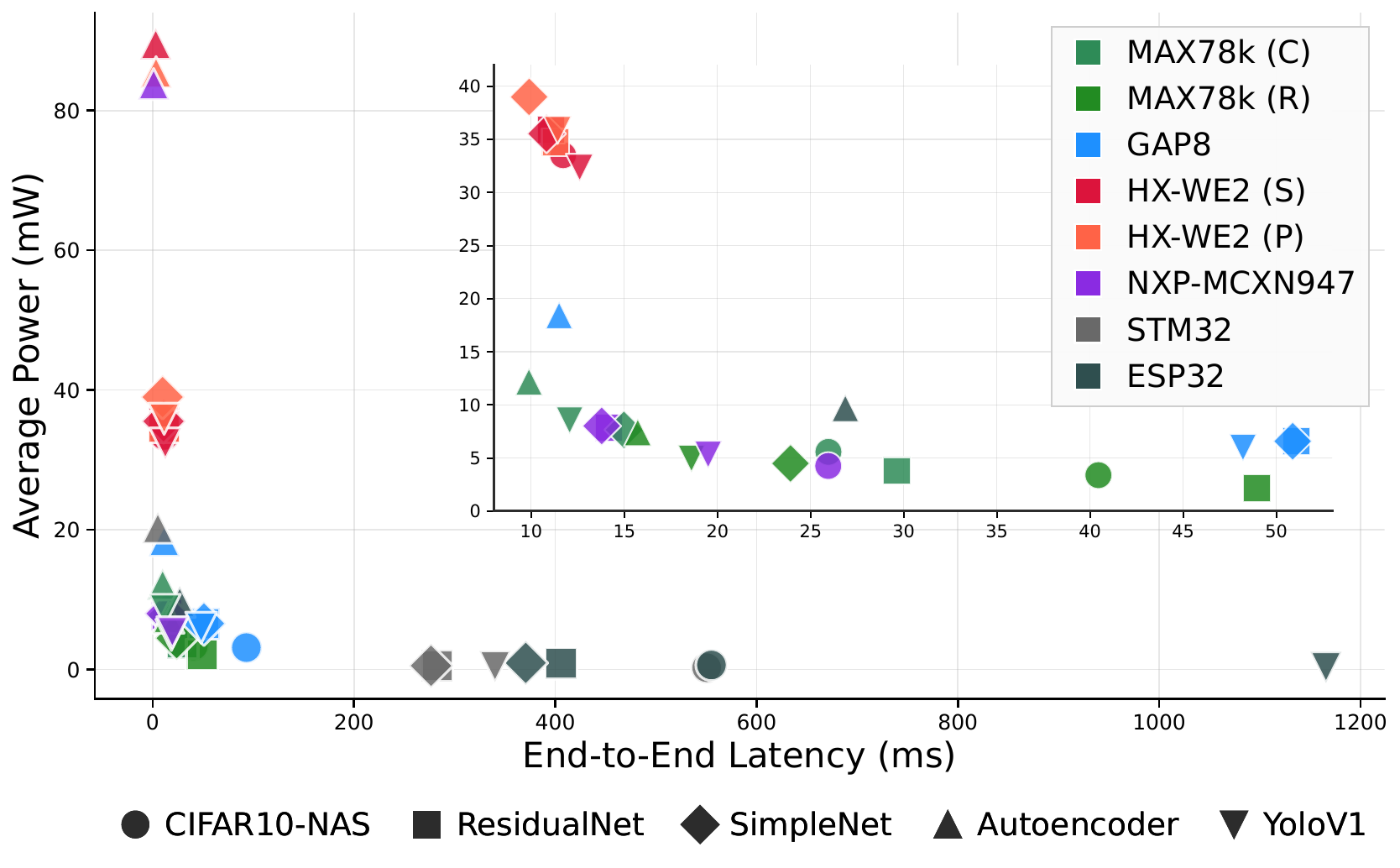}
\vspace{-0.6cm}
\caption{a visualization of average end-to-end latency \textit{vs.} power draw for evaluated models and platforms. The inset graph provides a magnified view of $\mu$NPU platforms with lower end-to-end latency.}\label{fig:power_vs_latency}
\vspace*{-1.2cm}
\end{figure}

\subsection{Latency and Memory I/O Breakdown }

\subsubsection{NPU initialization}

NPU initialization times vary significantly across the benchmark platforms, from as low as 0.07 ms on the MAX78000 to 12.94 ms on the GAP8.

However, the actual initialization overhead, with respect to end-to-end latency, is almost negligible on most $\mu$NPU platforms except the GAP8 (7.46 ms to 12.92 ms initialization latency across the various benchmark models). Such overhead could again be problematic for 
duty-cycled applications, where models must be frequently loaded/unloaded.

\subsubsection{Memory I/O}
Table \ref{tab:FlashRAM} details flash and RAM usage across our various benchmark platforms and models.

The significant memory I/O latency across all models on the MAX78000 forms an obvious inference bottleneck, with an average of 6.10x and 9.80x (Cortex-M4 and RISC-V) longer spent on memory I/O than actual inference ($e.g.$, 44.89 ms vs. 2.96 ms for ResidualNet with the MAX78k (R), meaning over 90\% of end-to-end inference time is dedicated to memory operations rather than computation). This implies the MAX78000's performance is largely memory-bound, 
and aligns with previous standalone benchmarks \cite{moss2022ultra}. Notably, memory I/O operations are more efficient on the MAX78000’s Cortex-M4 CPU than its RISC-V one. In contrast, memory I/O operations introduce negligible overhead across the other benchmark platforms with shared SRAM. 

\begin{table*}[htbp]
\centering
\caption{Flash and RAM use (KB) for evaluated models and platforms. 
The model with highest flash/RAM for each platform is \textbf{bolded}. 
Note: MAX78k's RAM is split into CPU-only and NPU-only.}
\label{tab:FlashRAM}
\vspace{-0.3cm}
\scriptsize
\renewcommand{\arraystretch}{1.15}
\resizebox{\textwidth}{!}{%
\begin{tabular}{lcccccccccccccccc}
\toprule
& \multicolumn{2}{c}{\textbf{MAX78k (C)}} 
& \multicolumn{2}{c}{\textbf{MAX78k (R)}} 
& \multicolumn{2}{c}{\textbf{GAP8}} 
& \multicolumn{2}{c}{\textbf{NXP-MCXN947}} 
& \multicolumn{2}{c}{\textbf{HX-WE2 (S)}} 
& \multicolumn{2}{c}{\textbf{HX-WE2 (P)}} 
& \multicolumn{2}{c}{\textbf{STM32H7A3}} 
& \multicolumn{2}{c}{\textbf{ESP32s3}} \\
\cmidrule(lr){2-3} \cmidrule(lr){4-5} \cmidrule(lr){6-7} \cmidrule(lr){8-9} 
\cmidrule(lr){10-11} \cmidrule(lr){12-13} \cmidrule(lr){14-15} \cmidrule(lr){16-17}
& \textbf{Flash} & \textbf{RAM} 
& \textbf{Flash} & \textbf{RAM} 
& \textbf{Flash} & \textbf{RAM} 
& \textbf{Flash} & \textbf{RAM} 
& \textbf{Flash} & \textbf{RAM} 
& \textbf{Flash} & \textbf{RAM} 
& \textbf{Flash} & \textbf{RAM} 
& \textbf{Flash} & \textbf{RAM} \\
\midrule
\textbf{NAS} &
347.67 & 4.96+295.51 &
364.39 & 6.16+295.51 &
\textbf{358.46} & \textbf{534.56} &
\textbf{569.94} & 371.70 &
127.75 & 551.87 &
127.75 & 538.59 &
423.61 & 93.75 &
674.57 & 268.86 \\
\textbf{ResNet} &
\textbf{425.38} & \textbf{4.98+372.84} &
\textbf{446.92} & \textbf{6.91+372.84} &
258.32 & 372.49 &
471.52 & 381.89 &
127.75 & \textbf{618.11} &
127.75 & \textbf{694.33} &
\textbf{456.07} & 70.97 &
694.44 & 268.78 \\
\textbf{SimpleNet} &
214.61 & 5.00+162.55 &
233.04 & 6.87+162.55 &
258.26 & 351.21 &
471.08 & 381.90 &
127.75 & 553.18 &
127.73 & 566.67 &
451.86 & 53.48 &
\textbf{698.06} & 268.77 \\
\textbf{Autoenc} &
184.15 & 6.46+133.59 &
193.74 & 6.09+133.59 &
143.31 & 196.20 &
261.36 & 381.27 &
125.44 & 336.06 &
125.44 & 336.35 &
203.57 & 21.35 &
445.59 & \textbf{271.89} \\
\textbf{YoloV1} &
130.43 & 6.93+41.75 &
147.96 & 8.38+41.75 &
43.29 & 159.46 &
287.70 & \textbf{410.83} &
\textbf{152.32} & 263.81 &
\textbf{152.32} & 319.10 &
119.28 & \textbf{167.52} &
355.19 & 268.77 \\
\bottomrule
\end{tabular}%
}
\end{table*}

Differing from CPUs and GPUs, which rely on a 1D contiguous memory space, $\mu$NPU hardware adopts a 2D memory layout; in this layout, one axis maps to parallel compute cores and the other organizes the logical address space. As shown in Fig. \ref{fig:arch}, each PE is equipped with its own weight memory space to avoid memory contention and maximize parallelization. This results in a hierarchical architecture with both channel-wise and weight-wise parallelism, though with the constraint that weights must use the same offset.

Recent work \cite{tinymem} has explored optimizing weight loading strategies for such 2D memory layouts to shrink I/O latency when switching models on a single device, including virtualizing weight memory within the accelerator to reduce fragmentation, optimizing dynamic weight allocation to minimize loading/unloading overhead,and weight preloading, where the next model’s weights are loaded by the idle CPU into unused memory regions before execution.

Further work should include automating memory management, alongside reducing I/O latency for single-model execution, using techniques like just-in-time prefetching, dynamic quantization, or input-adaptive pruning.



\subsubsection{Inference}
Another unexpected finding is the superior inference latency of the MAX78000 compared to the HX-WE2. The MAX78000 (C), for example, demonstrates an average $\sim$2.48× latency improvement over the HX-WE2 (P), despite the HX-WE2’s much higher advertised peak compute capacity at its maximum rated frequency (512 GOP/s at 1 GHz vs. 30 GOP/s for the MAX78000). In our benchmark, however, both platforms are operated at 100 MHz, so these peak figures are not directly comparable. The observed advantage may be attributed to more optimized weight-stationary dataflow patterns for CNN workloads compared to the Arm Ethos-U55. However, the HX-WE2 still wins in terms of end-to-end latency with much reduced memory I/O latency. The relatively consistent inference times across different models on the HX platforms also suggest its architecture is optimized for larger models than those in our benchmark suite. The MAX78000 demonstrates more variability in inference latency (ranging from 0.14 ms to 4.63 ms), suggesting greater scalability across differing model complexities.

The GAP8 demonstrates the highest end-to-end latency across all models - averaging 17× slower than the MAX78000, despite having similar compute capacity (22.65 GOPs vs. 30 GOPs on the MAX78000). However, again, the GAP8's large flash and RAM size make it more suitable for deploying large models or MoE architectures

Architectural factors help explain scaling patterns. Models like CIFAR10-NAS, generated via OFA NAS, make heavy use of frequent 1×1 convolutions and irregular channel scaling — operations that can map efficiently to NPUs with optimized channel-mixing kernels, but which can be memory-bound if channel-width transitions cause repeated buffer reallocations. ResidualNet and SimpleNet share identical convolutional footprints and late-stage expansion to 512 channels, but ResidualNet’s skip connections extend activation lifetimes and reduce buffer reuse, often increasing both memory pressure and execution time on platforms without aggressive activation scheduling. 


\subsubsection{CPU Post-Processing}
Post-processing operations, while often overlooked in benchmarking studies, can contribute to end-to-end latency and overall efficiency. We find CPU processing overhead is generally low across most of the evaluated platforms, in comparison to other execution stages, but is non-negligible for YoloV1’s NMS on certain platforms. For instance, the MAX78000 with RISC-V CPU active takes 3.82 ms in post-processing for YoloV1, compared to 2.62 ms spent in actual inference. This outlines the importance of minimizing CPU-dependent post-processing, and highlights a key design consideration with our benchmark; by ensuring all models are fully NPU-compatible across the various platforms, we aim to enable a fair comparison of end-to-end latency, avoiding bottlenecks or penalties caused by unsupported operators falling back to CPU execution. However, in real-world use, developers would build models that are optimized for a given target platform, making it necessary to consider the range of supported operators (which is quite limited on certain NPUs), and accuracy or performance trade-offs that might arise from using other, more compute-capable platforms, with more complex or unmodified models. 

\subsection{Task-Specific Considerations}


\noindent\textbf{Memory Constraints and Model Complexity}
Memory capacity significantly influences the feasible model complexity for each platform. The GAP8's expansive memory (8MB RAM, 20MB flash) enables deployment of substantially larger models than possible on the MAX78000 (512KB NPU memory, 128KB CPU memory), for example. This difference becomes critical for applications requiring more complex models, such as multi-class object detection or audio classification with large vocabulary sets.

The detailed memory I/O timing data provides additional insights into how different platforms handle model loading. The MAX78000's long memory I/O times (8.84 - 26.53 ms) are more suitable for persistent model deployment. In contrast the HX-WE2's comparatively large flash memory and low-latency memory I/O (0.03 - 1.11 ms), but longer initialization times (2.56 - 2.60 ms), are ideal for continuous inference or dynamic model switching.

Peak activation RAM and lifetime pressure metrics (Table~\ref{tab:model_intro}) highlight why certain models stress specific platforms more than others. For example, CIFAR10-NAS’s high lifetime pressure (18.39 MB$\cdot$steps) results in staggered channel scaling and interleaved pooling, which hold intermediate activations live for long durations. This amplifies SRAM demand and can degrade throughput on platforms with smaller or rigidly partitioned on-chip memories. 

\noindent\textbf{Operational Modes and Power Profiles}
The ability to support different operational modes significantly impacts a platform's suitability for specific applications. 
The MAX78000 displays high power variation between idle (10.87 - 13.21 mW) and inference (21.13 - 81.67 mW) states; hence, power gating -- shutting down unused compute domains or memory banks -- could extend its battery life in duty-cycled or event-driven inference scenarios. Moreover, dual-CPU platforms with asymmetric co-processing capabilities could improve task distribution between cores -- or enable hierarchical wake-up of CPU cores -- leading to power-saving advantages. For instance, MAX78000’s combination of RISC-V and Cortex-M4 cores, when used in tandem alongside early-exit strategies for dynamic, low-power inference, could further optimize energy usage during model deployment.

Our measurements suggest that platforms with large differences between idle and active power (e.g., MAX78000) stand to benefit more from aggressive gating than those with high static draw (e.g., HX-WE2).

\noindent\textbf{Precision Requirements and Quantization Support}
The bit-width support of each platform represents another important consideration for application-specific deployment. The MAX78000's support for 1, 2, 4, and 8-bit operations enables highly optimized model deployment for applications where lower precision is acceptable, or for models amenable to aggressive quantization.

Conversely, applications requiring higher numerical precision may benefit from platforms like the HW-WE2, which supports floating-point acceleration up to 32-bit precision. 

\subsection{Summary of Results}
We measured power consumption and latency for various model architectures across commercially-available $\mu$NPU platforms. We find GOPS isn't a reliable predictor for estimating end-to-end latency, and memory bandwidth enormously impacts performance. The architectural footprint of each model modulates its real-world performance. Memory-bound designs with high activation lifetime ($e.g.$, ResidualNet, CIFAR10-NAS) tend to underutilize available GOPS, whereas depth-heavy but narrow networks ($e.g.$, YOLOv1) appear to approach peak throughput on accelerators with low memory I/O cost. Models with limited convolutions see diminishing returns from NPU offload, explaining cases where MCUs perform competitively.

The MAX78000 $\mu$NPU, with its Cortex-M4 CPU active, offers the best \textit{overall} efficiency, with NPU initialization considered, delivering consistent sub-30ms end-to-end latency across all models. However, its performance is primarily memory-bound, spending up to 90\% of execution time on memory I/O operations. The HX-WE2 platform achieves an average end-to-end $\sim$1.93x speedup over MAX78000 but with $\sim$3.13x higher power consumption. The NXP-MCXN947 also offers relatively comparable (<30ms) end-to-end latency, with fast initialization and memory I/O; despite its lower computational throughput, it exhibits high efficiency on our lower-complexity, memory-light benchmark models. 

General-purpose MCUs demonstrate significantly lower efficiency, empirically validating the advantage of having dedicated neural hardware. 

Excluding initialization overhead ($i.e.$, for applications requiring continuous operation), the MILK-V ranks overall highest in terms efficiency, with its large idle power draw outweighed by fast end-to-end inference latency. 


\subsection{Future Directions}

\noindent\textbf{Advancing Hardware Architectures}: Developing next-generation $\mu$NPU architectures with larger on-chip cache and improved memory throughput is an obvious priority. This would (1) reduce the significant memory I/O overheads observed in certain platforms ($e.g.$, MAX78000) and (2) enable deployment of larger, more capable models or MoE architectures for context-aware inference.

\noindent\textbf{Optimizing Model Weight-Loading:} Together with hardware advancements, improved optimization of model architectures and loading strategies to maximize data reuse is also essential. The substantial memory I/O bottlenecks observed across certain platforms underscore the need for $\mu$NPU-specialized weight virtualization, dynamic allocation optimization, and prefetching strategies.

\noindent\textbf{Expanding Operator Support:} Currently, most $\mu$NPU platforms exhibit very limited operator support, heavily optimized for convolutional layers and a small set of accompanying functions such as pooling, elementwise activation, and basic addition. This specialization allows aggressive hardware acceleration for CNN-based workloads, but it comes at the cost of excluding operators central to other model architectures. For example, transformer-based architectures require efficient support for dense matrix multiplication (with high-rank tensors), softmax, normalization layers, and attention-specific dataflows -- none of which are implemented in $\mu$NPU instruction sets. 
In principle, unsupported layers could be offloaded to the CPU, 
but such heterogeneous execution requires a tightly integrated CPU/NPU software stack capable of minimizing data transfer and synchronization overheads. Notably, this kind of fine-grained CPU offloading is not supported on most of the evaluated platforms; 
as a result, our benchmarking is necessarily constrained to CNN variants that map well onto the hardware primitives.

\noindent\textbf{Improving Quantization and Model Compression:} Fine-grained bit-width quantization and other non-standard model optimizations also remain inadequately supported across $\mu$NPU platforms. This includes both a hardware and a software aspect, with existing software libraries designed for NN models on resource-constrained devices also generally lacking flexibility; TFLite/LiteRT, for example, only supports 8-bit integer and 16-bit float weight quantization.

\noindent\textbf{Enabling On-Device Training:}
Current $\mu$NPU architectures are optimized exclusively for inference and lack any hardware or software provisions for on-device training. 
This would enable privacy-preserving domain adaptation in \\bandwidth- or connectivity-constrained environments. This will require both architectural support for backpropagation alongside memory-efficient training algorithms capable of operating within the stringent compute, storage, and energy constraints inherent to $\mu$NPU-class hardware.

\noindent\textbf{Standardizing Model Formats:} The heterogeneity in supported model formats across our various benchmark platforms is another issue. Vendors should aim to move towards unified model formats to 
reduce cross-platform 
overheads.

\noindent\textbf{Developing Accurate Simulators:} Finally, reliable software simulators and predictive models for inference latency, power consumption, and memory utilization are notably absent for $\mu$NPUs (and MCUs in general). Such tools would enable developers to optimize deployments without physical hardware, accelerating the end-to-end development cycle. 

\subsection{Practical Recommendations}
We offer the following practical recommendations to embedded developers and hardware designers:

\noindent\textbf{For Energy-Efficiency}: The MAX78000  largely outperforms other $\mu$NPU platforms in terms of energy-efficiency (when including NPU initialization overhead), making it particularly well-suited for low- and battery-powered applications. For extended battery life, consider leveraging its ability to power-gate portions of the system during idle periods.

\noindent\textbf{For Latency-Critical Applications}: The HX-WE2 platform offers low-latency with rapid NPU initialization, memory I/O, and inference itself, making it best suited for applications requiring responsive model switching, real-time adaptation to changing conditions, or intermittent/duty-cycled operation. The NXP-MCXN947 also achieves relatively low end-to-end inference latency, at a significantly lower power budget, making it ideal for power-constrained workloads. Meanwhile, for latency-critical but space-constrained applications -- where power consumption is less of a concern and workloads avoid frequent NPU initialization -- developers should explore SoC architectures. Further evaluation is needed for NPU-equipped SoC platforms \cite{luckfox, k230}.

\noindent\textbf{For Large Models:} The GAP8's expansive memory makes it uniquely suitable for deploying larger models, or model-switching approaches where multiple specialized NNs are employed based on operating conditions (despite its longer initialization times and inference latency). However, again, if power consumption isn't a major concern, SoC-type platforms, with their low inference latency and larger memory capacity, could be a strong alternative.

\noindent\textbf{For Conv-Lite Models:} For models with limited convolutional layers, general-purpose MCUs can achieve competitive efficiency without dedicated neural acceleration, potentially eliminating the need for specialized hardware.


\subsection{Limitations}


\noindent{\textbf{Frequency Standardization:}} While enforcing a uniform CPU frequency across all platforms enables direct comparison of architectural efficiencies, it fails to showcase each platform's peak performance -- for example, many of the benchmark platforms can operate at higher frequencies than evaluated. Many platforms also combine DVFS with selective domain gating, so the net benefit of frequency scaling depends on both the workload and the platform.

\noindent{\textbf{Fixed Quantization Bit-Width:}} While the fixed-\texttt{INT8} quantization approach enables cross-platform fairness, it also masks important trade-offs. Certain $\mu$NPUs, such as the MAX78000, can gain substantial latency and energy savings from ultra-low bit-width quantization (1–4 bits), whereas others, like the HX-WE2, can preserve accuracy for sensitive workloads by exploiting \texttt{FLOAT16} or \texttt{FLOAT32} acceleration. 

\noindent{\textbf{CPU Configuration:}} We also enforced uniform CPU divider settings across experiments; however, many platforms support variable divider configurations, which could  potentially impact overall efficiency profiles.

\noindent{\textbf{Model Adaptation Constraints:}} The requirement to maintain structural consistency across all platforms necessitated compromises in model optimization. Platform-specific optimizations might yield slightly different efficiency profiles than our standardized approach.

\noindent{\textbf{Operator Support:}} Similarly, by ensuring all models are fully NPU-compatible across the various evaluated platforms, we negate the impact of unsupported NN operators. Further work should examine performance scaling across platforms with different sets of supported operators, using more complex or unmodified models, alongside precision-optimized models for each platform, and the impact of platform-specific architectural optimizations.



\section{RELATED WORK}
\noindent\textbf{Benchmarking NN Models on Constrained Hardware:}
Japana et al.'s MLPerf benchmark introduced the first industry-standard open-source framework for performance evaluation of NNs on mobile devices equipped with diverse NN accelerators and software stacks \cite{janapa2022mlperf}. Laskaridis et al. recently investigated the efficiency of large language models (LLMs) on various SOTA mobile platforms, including Android, iOS and Nvidia Jetson devices \cite{laskaridis2024melting}. Reuther et al. explored the performance and power characteristics of a wide range of NN accelerators, spanning cellular GPUs, FPGA accelerators, up to data center hardware \cite{reuther2019survey}. However, existing work on $\mu$NPU platforms has been limited to application-level performance assessments \cite{wang2022bed, song2024tada} or single-platform standalone benchmarks \cite{moss2022ultra, huang2024energy}. 

\noindent\textbf{NN Accelerators for MCUs:} 
NN accelerators offer vast potential in mitigating the computational and memory bottlenecks of traditional MCUs for NN inference. Beyond commercial accelerators ($e.g.$, Arm Ethos-U55), recent work has introduced new, more efficient custom designs. For instance, Venkataramani et al. designed RaPiD, an accelerator tailored for ultra-low-power INT4 inference, achieving an energy efficiency of 3-13.5 TOPS/W (average 7 TOPS/W) \cite{venkataramani2021rapid}. Conti et al. developed the XNOR Neural Engine, a digital, configurable hardware accelerator IP for binary neural networks, integrated into an MCU with an autonomous I/O subsystem and hybrid SRAM/standard cell memory \cite{conti2018xnor}.

\noindent\textbf{Efficient On-Device Inference:}
Numerous works have explored model compression \cite{choudhary2020comprehensive, lin2020mcunet, zawish2024complexity}, the design of more efficient NN operators/architectures for lower resource usage \cite{pan2023reusing,tarnawski2020efficient,li2017simple}, and adaptive NN inference based on input complexity and workload \cite{ee, delight, hydranets}. Various hardware-based optimizations have also been studied, such as parallel dataflow processing \cite{gong2025dex}. Our work aims to further advance efficient NN deployment across $\mu$NPU platforms by identifying current hardware bottlenecks.

\section{CONCLUSION}

Our evaluation of NN models across commercially-available $\mu$NPUs reveals both expected trends and surprising findings. We show that dedicated neural accelerators deliver up to two orders of magnitude higher energy-efficiency than MCUs, but that theoretical capacity ($i.e.$, GOPs) alone poorly predicts real-world performance. Our stage-by-stage breakdown reveals critical bottlenecks on certain platforms -- particularly in memory I/O operations -- alongside key insights for future work in hardware and model design. We encourage developers to consider trade-offs in latency, energy-efficiency, model complexity, and flexibility for optimal deployment. We open-source our benchmarking toolchain and hope its use can streamline cross-platform compilation and evaluation.

\section{ACKNOWLEDGMENTS}
This research was supported in part by the UKRI Open Plus Fellowship (EP/W005271/1: Securing the Next Billion Consumer Devices on the Edge), as well as funding from the Grantham Institute, Imperial College London.

\bibliographystyle{unsrt}
\bibliography{sample-base}

\newpage
\clearpage
\onecolumn

\section*{Supplementary Material}

\captionof{table}{Complete latency (ms) and power (mW) measurements across each stage, model, and platform.}
\begin{center}
\rotatebox{90}{%
\label{tab:mega-comparison}
\resizebox{\textwidth}{!}{%
\begin{tabular}{ll*{8}{cc}}
\toprule
\textbf{Model} & \textbf{Stage} 
& \multicolumn{2}{c}{\textbf{MAX78k (C)}} 
& \multicolumn{2}{c}{\textbf{MAX78k (R)}} 
& \multicolumn{2}{c}{\textbf{GAP8}} 
& \multicolumn{2}{c}{\textbf{NXP-MCXN947}} 
& \multicolumn{2}{c}{\textbf{HX-WE2 (S)}} 
& \multicolumn{2}{c}{\textbf{HX-WE2 (P)}} 
& \multicolumn{2}{c}{\textbf{STM32H7A3ZI}} 
& \multicolumn{2}{c}{\textbf{ESP32}} \\
\cmidrule(lr){3-4}\cmidrule(lr){5-6}\cmidrule(lr){7-8}\cmidrule(lr){9-10}%
\cmidrule(lr){11-12}\cmidrule(lr){13-14}\cmidrule(lr){15-16}\cmidrule(lr){17-18}
& & \textbf{Time} & \textbf{Power} & \textbf{Time} & \textbf{Power} & \textbf{Time} & \textbf{Power}
& \textbf{Time} & \textbf{Power} & \textbf{Time} & \textbf{Power} & \textbf{Time} & \textbf{Power}
& \textbf{Time} & \textbf{Power} & \textbf{Time} & \textbf{Power} \\
\midrule
\multirow{5}{*}{\textbf{NAS}} 
& Initialization    & 0.074{\scriptsize ±0.001} & 13.67{\scriptsize ±0.02} & 0.169{\scriptsize ±0.002} & 12.84{\scriptsize ±0.03} & 12.939{\scriptsize ±0.237} & 102.93{\scriptsize ±3.11} & 0.219{\scriptsize ±0.001} & 112.29{\scriptsize ±3.39} & 2.597{\scriptsize ±0.002} & 91.50{\scriptsize ±0.22} & 2.601{\scriptsize ±0.001} & 91.44{\scriptsize ±0.19} & 0.429{\scriptsize ±0.005} & 47.38{\scriptsize ±0.07} & 86.727{\scriptsize ±0.001} & 105.93{\scriptsize ±0.09} \\
& Memory I/O        & 21.241{\scriptsize ±0.004} & 25.05{\scriptsize ±0.05} & 35.529{\scriptsize ±0.003} & 22.66{\scriptsize ±0.03} & 2.661{\scriptsize ±0.007} & 38.58{\scriptsize ±0.95} & 0.049{\scriptsize ±0.001} & 117.26{\scriptsize ±0.84} & 0.123{\scriptsize ±0.001} & 97.55{\scriptsize ±1.25} & 0.123{\scriptsize ±0.001} & 97.01{\scriptsize ±0.71} & — & — & — & — \\
& Inference         & 4.625{\scriptsize ±0.002}  & 80.41{\scriptsize ±0.08} & 4.642{\scriptsize ±0.002}  & 79.03{\scriptsize ±0.05} & 77.396{\scriptsize ±0.470} & 117.22{\scriptsize ±1.47} & 27.267{\scriptsize ±0.002} & 118.03{\scriptsize ±0.65} & 8.988{\scriptsize ±0.022} & 112.35{\scriptsize ±0.76} & 8.319{\scriptsize ±0.008} & 114.62{\scriptsize ±0.76} & 550.010{\scriptsize ±0.012} & 54.91{\scriptsize ±0.09} & 468.381{\scriptsize ±0.001} & 151.99{\scriptsize ±0.31} \\
& Post-Processing   & 0.016{\scriptsize ±0.001}  & 25.66{\scriptsize ±0.04} & 0.111{\scriptsize ±0.001}  & 22.43{\scriptsize ±0.06} & 0.076{\scriptsize ±0.001}   & 33.23{\scriptsize ±1.28} & 0.010{\scriptsize ±0.001} & 109.62{\scriptsize ±1.19} & 0.007{\scriptsize ±0.001} & 91.19{\scriptsize ±1.31} & 0.007{\scriptsize ±0.001} & 91.17{\scriptsize ±1.39} & 0.015{\scriptsize ±0.001}  & 46.01{\scriptsize ±0.05} & 0.017{\scriptsize ±0.001}  & 101.93{\scriptsize ±1.86} \\
& Idle              & —                         & 13.21{\scriptsize ±0.02} & —                         & 10.87{\scriptsize ±0.01} & —                          & 33.67{\scriptsize ±0.35} & —                         & 105.71{\scriptsize ±0.14} & —                         & 89.09{\scriptsize ±1.30} & —                         & 89.09{\scriptsize ±1.30} & —                          & 38.13{\scriptsize ±0.05} & —                          & 77.73{\scriptsize ±0.07} \\
\addlinespace
\multirow{5}{*}{\textbf{ResNet}} 
& Initialization    & 0.074{\scriptsize ±0.001} & 13.87{\scriptsize ±0.05} & 0.169{\scriptsize ±0.001} & 12.82{\scriptsize ±0.07} & 10.133{\scriptsize ±0.001} & 145.90{\scriptsize ±3.59} & 0.221{\scriptsize ±0.001} & 110.37{\scriptsize ±1.99} & 2.593{\scriptsize ±0.001} & 94.27{\scriptsize ±0.23} & 2.602{\scriptsize ±0.001} & 93.51{\scriptsize ±2.01} & 0.432{\scriptsize ±0.002} & 48.32{\scriptsize ±0.07} & 87.489{\scriptsize ±0.059} & 106.41{\scriptsize ±0.07} \\
& Memory I/O        & 26.526{\scriptsize ±0.002} & 25.14{\scriptsize ±0.05} & 44.868{\scriptsize ±0.003} & 22.64{\scriptsize ±0.04} & 2.471{\scriptsize ±0.110} & 38.39{\scriptsize ±0.31} & 0.050{\scriptsize ±0.001} & 115.95{\scriptsize ±0.99} & 0.123{\scriptsize ±0.001} & 99.15{\scriptsize ±0.45} & 0.136{\scriptsize ±0.001} & 97.74{\scriptsize ±1.37} & — & — & — & — \\
& Inference         & 2.893{\scriptsize ±0.001}  & 47.12{\scriptsize ±0.09} & 2.964{\scriptsize ±0.001}  & 46.86{\scriptsize ±0.07} & 38.447{\scriptsize ±0.081} & 113.47{\scriptsize ±0.40} & 16.067{\scriptsize ±0.002} & 118.57{\scriptsize ±0.52} & 8.295{\scriptsize ±0.001} & 110.13{\scriptsize ±0.79} & 8.535{\scriptsize ±0.015} & 109.71{\scriptsize ±2.35} & 282.000{\scriptsize ±0.010} & 54.98{\scriptsize ±0.08} & 318.149{\scriptsize ±0.001} & 144.91{\scriptsize ±0.27} \\
& Post-Processing   & 0.123{\scriptsize ±0.001}  & 26.10{\scriptsize ±0.06} & 0.963{\scriptsize ±0.002}  & 22.45{\scriptsize ±0.05} & 0.037{\scriptsize ±0.001}   & 37.41{\scriptsize ±1.55} & 0.039{\scriptsize ±0.001} & 109.55{\scriptsize ±1.95} & 0.038{\scriptsize ±0.001} & 93.19{\scriptsize ±0.15} & 0.038{\scriptsize ±0.001} & 91.95{\scriptsize ±1.27} & 0.082{\scriptsize ±0.002}  & 50.47{\scriptsize ±0.09} & 0.052{\scriptsize ±0.001}  & 102.79{\scriptsize ±3.71} \\
& Idle              & —                         & 13.21{\scriptsize ±0.02} & —                         & 10.87{\scriptsize ±0.01} & —                          & 33.67{\scriptsize ±0.35} & —                         & 105.71{\scriptsize ±0.14} & —                         & 89.09{\scriptsize ±1.30} & —                         & 89.09{\scriptsize ±1.30} & —                          & 38.13{\scriptsize ±0.05} & —                          & 77.73{\scriptsize ±0.07} \\
\addlinespace
\multirow{5}{*}{\textbf{SimpleNet}} 
& Initialization    & 0.074{\scriptsize ±0.001} & 13.71{\scriptsize ±0.04} & 0.168{\scriptsize ±0.002} & 12.81{\scriptsize ±0.03} & 10.017{\scriptsize ±0.002} & 137.44{\scriptsize ±3.95} & 0.221{\scriptsize ±0.001} & 108.98{\scriptsize ±0.42} & 2.560{\scriptsize ±0.003} & 95.75{\scriptsize ±0.49} & 2.597{\scriptsize ±0.001} & 91.51{\scriptsize ±0.21} & 0.431{\scriptsize ±0.001} & 48.13{\scriptsize ±0.04} & 87.372{\scriptsize ±0.001} & 106.58{\scriptsize ±0.66} \\
& Memory I/O        & 12.137{\scriptsize ±0.003} & 24.46{\scriptsize ±0.07} & 20.220{\scriptsize ±0.003} & 22.27{\scriptsize ±0.04} & 2.598{\scriptsize ±0.076} & 38.51{\scriptsize ±0.69} & 0.049{\scriptsize ±0.001} & 117.54{\scriptsize ±1.69} & 0.123{\scriptsize ±0.001} & 91.11{\scriptsize ±1.77} & 0.123{\scriptsize ±0.001} & 96.97{\scriptsize ±0.69} & — & — & — & — \\
& Inference         & 2.657{\scriptsize ±0.002}  & 50.89{\scriptsize ±0.04} & 2.670{\scriptsize ±0.003}  & 49.86{\scriptsize ±0.05} & 38.226{\scriptsize ±0.117} & 122.12{\scriptsize ±3.70} & 13.523{\scriptsize ±0.001} & 118.40{\scriptsize ±0.81} & 8.099{\scriptsize ±0.001} & 107.15{\scriptsize ±0.69} & 7.134{\scriptsize ±0.001} & 106.84{\scriptsize ±0.77} & 276.000{\scriptsize ±0.011} & 55.19{\scriptsize ±0.06} & 283.210{\scriptsize ±0.002} & 139.96{\scriptsize ±1.84} \\
& Post-Processing   & 0.114{\scriptsize ±0.001}  & 25.62{\scriptsize ±0.07} & 0.864{\scriptsize ±0.001}  & 22.51{\scriptsize ±0.05} & 0.037{\scriptsize ±0.001}   & 36.83{\scriptsize ±0.39} & 0.038{\scriptsize ±0.001} & 109.33{\scriptsize ±0.21} & 0.039{\scriptsize ±0.001} & 90.54{\scriptsize ±0.14} & 0.038{\scriptsize ±0.001} & 90.47{\scriptsize ±0.21} & 0.102{\scriptsize ±0.002}  & 46.10{\scriptsize ±0.07} & 0.053{\scriptsize ±0.001}  & 105.11{\scriptsize ±3.25} \\
& Idle              & —                         & 13.21{\scriptsize ±0.02} & —                         & 10.87{\scriptsize ±0.01} & —                          & 33.67{\scriptsize ±0.35} & —                         & 105.71{\scriptsize ±0.14} & —                         & 89.09{\scriptsize ±1.30} & —                         & 89.09{\scriptsize ±1.30} & —                          & 38.13{\scriptsize ±0.05} & —                          & 77.73{\scriptsize ±0.07} \\
\addlinespace
\multirow{5}{*}{\textbf{AI8XAutoEnc}} 
& Initialization    & 0.074{\scriptsize ±0.001} & 13.75{\scriptsize ±0.06} & 0.168{\scriptsize ±0.001} & 12.86{\scriptsize ±0.05} & 8.174{\scriptsize ±0.076}  & 89.22{\scriptsize ±1.28} & 0.279{\scriptsize ±0.001} & 109.14{\scriptsize ±0.20} & 2.599{\scriptsize ±0.002} & 90.79{\scriptsize ±0.70} & 2.601{\scriptsize ±0.001} & 96.04{\scriptsize ±1.01} & 0.124{\scriptsize ±0.002} & 48.17{\scriptsize ±0.06} & 19.758{\scriptsize ±0.002} & 103.51{\scriptsize ±0.59} \\
& Memory I/O        & 9.663{\scriptsize ±0.001}  & 25.04{\scriptsize ±0.08} & 15.398{\scriptsize ±0.003} & 22.63{\scriptsize ±0.06} & 2.630{\scriptsize ±0.042}  & 38.55{\scriptsize ±0.56} & 0.150{\scriptsize ±0.001} & 112.87{\scriptsize ±1.21} & 0.031{\scriptsize ±0.001} & 92.75{\scriptsize ±1.83} & 0.031{\scriptsize ±0.001} & 94.03{\scriptsize ±0.62} & — & — & — & — \\
& Inference         & 0.143{\scriptsize ±0.001}  & 81.67{\scriptsize ±0.09} & 0.156{\scriptsize ±0.001}  & 81.34{\scriptsize ±0.07} & 0.696{\scriptsize ±0.015}  & 84.35{\scriptsize ±1.25} & 19.246{\scriptsize ±0.003} & 112.19{\scriptsize ±0.44} & 0.451{\scriptsize ±0.001} & 92.40{\scriptsize ±1.19} & 0.749{\scriptsize ±0.001} & 99.02{\scriptsize ±1.56} & 5.010{\scriptsize ±0.110} & 56.11{\scriptsize ±0.09} & 7.106{\scriptsize ±0.001}  & 157.17{\scriptsize ±0.17} \\
& Post-Processing   & —                         & —                        & —                         & —                        & —                          & —                        & —                         & —                        & —                         & —                        & —                         & —                        & —                         & —                        & —                         & — \\
& Idle              & —                         & 13.21{\scriptsize ±0.02} & —                         & 10.87{\scriptsize ±0.01} & —                          & 33.67{\scriptsize ±0.35} & —                         & 105.71{\scriptsize ±0.14} & —                         & 89.09{\scriptsize ±1.30} & —                         & 89.09{\scriptsize ±1.30} & —                          & 38.13{\scriptsize ±0.05} & —                          & 77.73{\scriptsize ±0.07} \\
\addlinespace
\multirow{5}{*}{\textbf{YoloV1}} 
& Initialization    & 0.074{\scriptsize ±0.001} & 13.90{\scriptsize ±0.03} & 0.169{\scriptsize ±0.001} & 12.87{\scriptsize ±0.02} & 7.457{\scriptsize ±0.001}  & 96.28{\scriptsize ±2.90} & 0.277{\scriptsize ±0.001} & 108.85{\scriptsize ±0.39} & 2.599{\scriptsize ±0.003} & 100.94{\scriptsize ±1.01} & 2.598{\scriptsize ±0.003} & 100.45{\scriptsize ±0.59} & 3.639{\scriptsize ±0.002} & 49.04{\scriptsize ±0.05} & 628.970{\scriptsize ±0.002} & 106.49{\scriptsize ±0.163} \\
& Memory I/O        & 8.841{\scriptsize ±0.003}  & 41.82{\scriptsize ±0.04} & 11.995{\scriptsize ±0.002} & 35.55{\scriptsize ±0.04} & 2.828{\scriptsize ±0.037}  & 38.10{\scriptsize ±0.11} & 0.514{\scriptsize ±0.001} & 110.51{\scriptsize ±1.61} & 1.106{\scriptsize ±0.001} & 100.31{\scriptsize ±3.09} & 1.106{\scriptsize ±0.001} & 101.95{\scriptsize ±2.90} & — & — & — & — \\
& Inference         & 2.612{\scriptsize ±0.002}  & 21.47{\scriptsize ±0.04} & 2.623{\scriptsize ±0.001}  & 21.13{\scriptsize ±0.05} & 36.109{\scriptsize ±0.047} & 113.67{\scriptsize ±2.39} & 19.251{\scriptsize ±0.001} & 113.31{\scriptsize ±1.66} & 8.477{\scriptsize ±0.001} & 112.63{\scriptsize ±1.30} & 7.295{\scriptsize ±0.001} & 111.69{\scriptsize ±0.97} & 336.000{\scriptsize ±0.013} & 54.99{\scriptsize ±0.08} & 536.222{\scriptsize ±0.001} & 129.74{\scriptsize ±0.107} \\
& Post-Processing   & 0.538{\scriptsize ±0.001}  & 25.87{\scriptsize ±0.06} & 3.818{\scriptsize ±0.001}  & 22.41{\scriptsize ±0.04} & 1.817{\scriptsize ±0.002}  & 39.89{\scriptsize ±0.62} & 0.112{\scriptsize ±0.001} & 109.35{\scriptsize ±0.15} & 0.419{\scriptsize ±0.001} & 93.63{\scriptsize ±1.45} & 0.420{\scriptsize ±0.001} & 95.65{\scriptsize ±0.79} & 0.476{\scriptsize ±0.001} & 45.71{\scriptsize ±0.04} & 0.534{\scriptsize ±0.001} & 109.57{\scriptsize ±0.08} \\
& Idle              & —                         & 13.21{\scriptsize ±0.02} & —                         & 10.87{\scriptsize ±0.01} & —                          & 33.67{\scriptsize ±0.35} & —                         & 105.71{\scriptsize ±0.14} & —                         & 89.09{\scriptsize ±1.30} & —                         & 89.09{\scriptsize ±1.30} & —                          & 38.13{\scriptsize ±0.05} & —                          & 77.736{\scriptsize ±0.07} \\
\bottomrule
\end{tabular}
}%
}
\end{center}

\raggedright
{\footnotesize\textbf{Notes}:\\
- For MCUs without neural hardware, STM32H7A3ZI and ESP32, Initialization and Memory I/O are combined.\\
- The post-processing for ResidualNet, SimpleNet, and NAS models is composed of a softmax operation.\\
- The post-processing for YOLOv1 is a NMS (non-max suppression) operation, also with softmax.}

\twocolumn

\end{document}